\documentclass[letterpaper]{article} 
\usepackage{aaai24}  
\usepackage{times}  
\usepackage{helvet}  
\usepackage{courier}  
\usepackage[hyphens]{url}  
\usepackage{graphicx} 
\usepackage{natbib}  
\usepackage{caption} 
\usepackage{xcolor}
\usepackage{multirow}
\definecolor{citecolor}{HTML}{0071bc}
\definecolor{ourscolor}{HTML}{c2d1e5}
\usepackage{colortbl}  
\usepackage{amssymb}
\usepackage{subcaption}
\usepackage{algorithm}
\usepackage{algorithmic}
\usepackage{amssymb}
\usepackage{pifont}

\usepackage{amsmath}
\usepackage{bm}
\newcommand{\cmark}{\ding{51}}%
\newcommand{\xmark}{\ding{55}}%

\newcommand{\ie}{\textit{i}.\textit{e}., }
\newcommand{\eg}{\textit{e}.\textit{g}., }%

\usepackage{newfloat}
\usepackage{listings}
\DeclareCaptionStyle{ruled}{labelfont=normalfont,labelsep=colon,strut=off} 
\floatstyle{ruled}
\newfloat{listing}{tb}{lst}{}
\floatname{listing}{Listing}
\nocopyright 

\title{CLIM: Contrastive Language-Image Mosaic for Region Representation}
\author{
    Size Wu\textsuperscript{\rm 1}, Wenwei Zhang\textsuperscript{\rm 1}, Lumin Xu\textsuperscript{\rm 2}
    \\ Sheng Jin\textsuperscript{\rm 3,4}, Wentao Liu\textsuperscript{\rm 4,5}, Chen Change Loy\textsuperscript{\rm 1}\thanks{Corresponding author.}
}
\affiliations{
    \textsuperscript{\rm 1}S-Lab, Nanyang Technological University,
    \textsuperscript{\rm 2}The Chinese University of Hong Kong \\
    \textsuperscript{\rm 3}The University of Hong Kong,
    \textsuperscript{\rm 4}SenseTime Research and Tetras.AI,
    \textsuperscript{\rm 5}Shanghai AI Laboratory \\

    \{size001, wenwei001, ccloy\}@ntu.edu.sg, 
    luminxu@link.cuhk.edu.hk,
    \{jinsheng, liuwentao\}@tetras.ai
}

\begin{document}

\maketitle
\begin{abstract} 
Detecting objects accurately from a large or open vocabulary necessitates the vision-language alignment on region representations.
However, learning such a region-text alignment by obtaining high-quality box annotations with text labels or descriptions is expensive and infeasible. In contrast, collecting image-text pairs is simpler but lacks precise object location information to associate regions with texts. 
In this paper, we propose a novel approach called Contrastive Language-Image Mosaic (CLIM), which leverages large-scale image-text pairs effectively for aligning region and text representations. 
CLIM combines multiple images into a mosaicked image and treats each image as a \emph{`pseudo region'}. The feature of each pseudo region is extracted and trained to be similar to the corresponding text embedding while dissimilar from others by a contrastive loss, enabling the model to learn the region-text alignment without costly box annotations.
As a generally applicable approach, CLIM consistently improves different open-vocabulary object detection methods that use caption supervision. Furthermore, CLIM can effectively enhance the region representation of vision-language models, thus providing stronger backbones for open-vocabulary object detectors. 
Our experimental results demonstrate that CLIM improves different baseline open-vocabulary object detectors by a large margin on both OV-COCO and OV-LVIS benchmarks. The code is available at \url{https://github.com/wusize/CLIM}.

\end{abstract}

\section{Introduction}\label{sec:introduction}

\begin{figure*}[ht]
\centering
\includegraphics[width=0.95\textwidth]{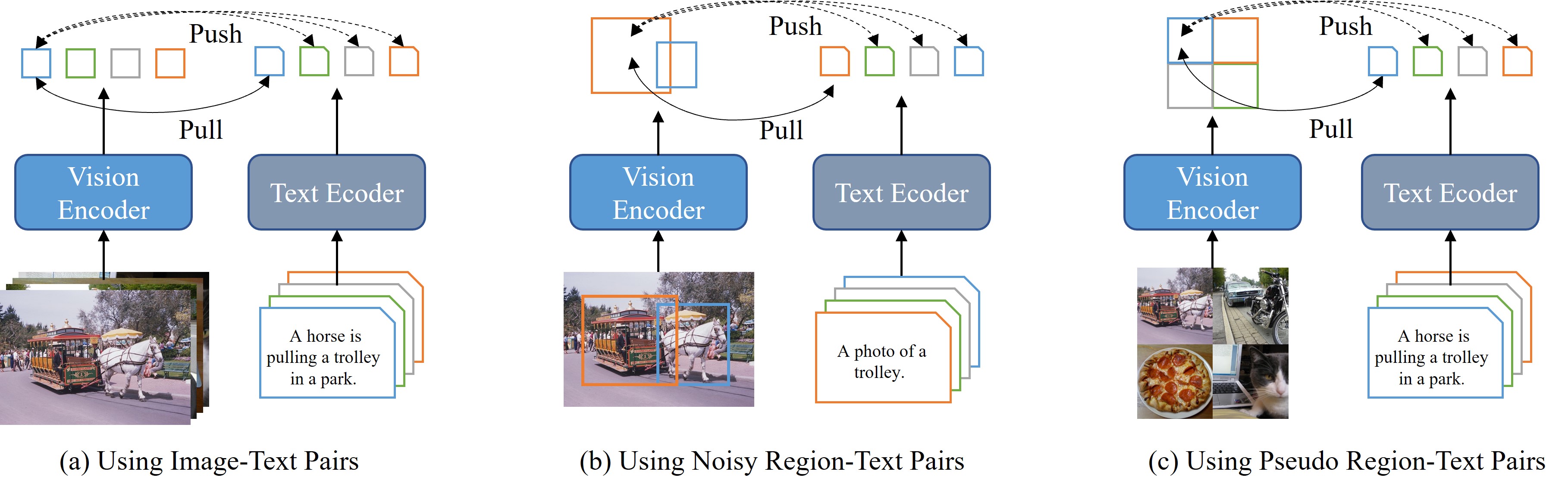}
\caption{Different strategies of learning vision-language alignment.
\textbf{(a)} Learning image-text alignment using image-text pairs.
\textbf{(b)} Learning region-text alignment using noisy and un-scalable region-text pairs.
\textbf{(c)} The proposed CLIM method mosaics images to generate pseudo region-text pairs for region-text alignment.
}
\label{fig:teaser}
\end{figure*}

Object detection is a fundamental task in computer vision that involves recognizing and localizing objects in the images. With the advent of deep learning, significant progress has been made in object detection on benchmark datasets that involve a confined set of categories, such as 80 classes in COCO~\cite{lin2014microsoft} and 20 classes in PASCAL VOC~\cite{everingham2010pascal}. However, to make object detection useful in real-world applications, it is essential to recognize objects in an open vocabulary that is inherently long-tailed and open-ended, with novel concepts that are not commonly seen in the benchmark datasets~\cite{oltr, WilliamJReed2001ThePZ}.

To obtain the generalization ability required by open-vocabulary recognition, modern open-vocabulary object detection methods either apply the image-text pairs as weak supervision to the training of object detection~\cite{zhou2022detecting} or reap the vision-language alignment from large-scale image-text pre-training~\cite{gu2021open, FVLM, wu2023baron}, which can be summarized as Figure~\ref{fig:teaser}(a). Due to lack of object location information, these methods fail to effectively transfer the vision-language alignment to region representations.

There are also attempts~\cite{zhong2022regionclip, VLDet} to learn vision-language alignment directly at the region level as shown in Figure~\ref{fig:teaser}(b). For example, RegionCLIP~\cite{zhong2022regionclip} matches region proposals with object nouns to generate region-text pairs. The region proposals are detected by a pre-trained Region Proposal Network (RPN) and the object nouns are obtained by parsing the image captions. However, the generated region-text pairs are inevitably noisy since both the localization of region proposals and the region-text matching can be inaccurate.

In this paper, we propose a novel approach to learn region-language alignment without the inaccurate and tedious region-text matching process. The proposed method, named Contrastive Language-Image Mosaic (CLIM), combines multiple images into a mosaicked image and conducts contrastive learning as shown in Figure~\ref{fig:teaser}(c). This process forces the representation of each sub-image in the mosaicked image to be similar to its corresponding text representation and dissimilar to the others. By treating the sub-images as \emph{`pseudo regions'}, CLIM facilitates the learning of region-text alignment while eliminating the need to annotate bounding boxes of objects.

CLIM prepares a canvas at each training iteration and divides it evenly into a flexible number of regions (\eg $2\times2$, $3\times3$ and $4\times4$). Each region is then filled by an image that is randomly sampled from the training dataset, termed as a \emph{`pseudo region'}. 
The canvas then becomes a mosaicked image, which is fed into the vision encoder as a whole to obtain a feature map. The feature of each pseudo region is extracted from the feature map using the corresponding box location in the mosaicked image.
Meanwhile, the texts of the original images are fed independently and parallelly to the text encoder. 
Finally, the features of pseudo regions are aligned with the text features in a contrastive manner.

We deliberately keep the design of CLIM simple so that it can be easily applied to different open-vocabulary object detection methods~\cite{zhou2022detecting, wu2023baron} as well as the pre-training of vision-language models~\cite{radford2021learning, zareian2021open}. To evaluate the effectiveness of CLIM, we test different design choices for both types of applications. When applied to the caption supervision branch of Detic~\cite{zhou2022detecting} and BARON~\cite{wu2023baron}, CLIM improves Detic by 5.1 AP$_{50}^{\mathrm{novel}}$ and BARON by 2.1 AP$_{50}^{\mathrm{novel}}$ on OV-COCO. On OV-LVIS, it improves Detic by 2.3 mAP$_{r}^{\mathrm{mask}}$. For the enhancement of CLIP models' region representation, the model trained by CLIM improves F-VLM~\cite{FVLM} by 2.2 mAP$_{r}^{\mathrm{mask}}$ on OV-LVIS. CLIM also boosts OV-RCNN~\cite{zareian2021open} on OV-COCO by 3.4 AP$_{50}^{\mathrm{novel}}$ when applied to OV-RCNN's vision-language pre-training stage.

\section{Related Work}\label{sec:related_work}
\paragraph{Learning Vision-Language Alignment.} 
Aligning visual and lingual representations is a key step to achieve general scene understanding~\cite{radford2021learning,jia2021scaling,zareian2021open,VILT, wu2023open}. CLIP models~\cite{radford2021learning} that are pre-trained on billion-scale image-text pairs have shown impressive zero-shot capabilities in downstream image recognition tasks. However, CLIP models lack awareness of local regions while building image-level alignment. To reason about image regions, RegionCLIP~\cite{zhong2022regionclip} fine-tunes CLIP models by exploiting the region-text correspondence between pseudo-labeled region proposals and object concepts. On the other hand, GLIP~\cite{li2021grounded} exploits visual-grounding datasets~\cite{visualgenome,GQA} that associate bounding boxes and text descriptions. Nonetheless, datasets with bounding box annotations needed by these methods are limited and expensive, while large-scale image-text pairs lack object location information, preventing more assertive exploration of vision-language alignment at the region level. In this paper, we circumvent this issue by mosaicking images and treating each image as a `pseudo region' for learning region-level representations using low-cost image-text annotations.

\paragraph{Weakly-Supervised Object Detection.}
The task of weakly-supervised object detection (WSOD) is to train detectors using image-level supervision. Some studies~\cite{li2019weakly,shen2019cyclic,wan2019c} rely on low-level region proposal techniques~\cite{uijlings2013selective, arbelaez2014multiscale} to localize objects. A more general form of WSOD, known as semi-supervised WSOD~\cite{mosaicos,ramanathan2020dlwl,redmon2017yolo9000}, allows the use of bounding box supervision together with image labels. In particular, MosaicOS~\cite{mosaicos} groups object-centric images into pseudo scene-centric images using mosaic augmentation to address the challenge of long-tail object detection. CLIM is similar to MosaicOS in mosaicking images. However, MosaicOS serves as an augmentation to re-balance the distributions of rare and frequent categories while CLIM aims at the representation learning for region-language alignment. Moreover, MosaicOS is specially designed for object detection while CLIM is a more generally applicable approach that applies to not only open-vocabulary detection methods but also vision-language pre-training.

\paragraph{Open-Vocabulary Object Detection.}
Open-vocabulary object detection (OVD)~\cite{zareian2021open} is concerned with detecting objects of novel categories that are unseen during training.
Some works utilize large pre-trained vision-language models (VLMs)~\cite{radford2021learning} to acquire open-vocabulary recognition ability by knowledge distillation~\cite{gu2021open,zang2022open, wu2023baron, wu2023clipself} or directly building open-vocabulary detectors upon the pre-trained VLMs~\cite{zareian2021open, FVLM, xu2023dst}. Others~\cite{gao2021towards, zhou2022detecting, wu2023baron} employ image-level supervision (\eg image captions) to learn a large number of novel concepts. In this paper, we first instantiate CLIM on the works that use image captions, \ie Detic~\cite{zhou2022detecting} and BARON~\cite{wu2023baron}.
Then, we apply CLIM to the vision-language pre-training so that the VLMs serve as stronger backbones to build open-vocabulary object detectors for F-VLM~\cite{FVLM} and OV-RCNN~\cite{zareian2021open}.

\section{Method}\label{sec:method}

In this section, we introduce CLIM that facilitates the learning of region-level visual-language alignment from image-text pairs, without relying on either costly bounding box annotations~\cite{GLIP} or inaccurate bounding box predictions~\cite{zhou2022detecting, VLDet}. 
Our method involves mosaicking images and treating each image as a \emph{`pseudo region'} in the context of the mosaicked image. The visual features of these pseudo regions are then extracted and aligned with their corresponding text features through contrastive learning. 
CLIM is versatile and can be applied to both open-vocabulary object detection methods and vision-language pre-training.

\subsection{Mosaicking Images as Pseudo Regions}\label{sec:mosaic_language_image}

Ideally, learning the alignment between region and text representations for generalizable object recognition would require one to annotate massive region-level bounding boxes that are labeled with text descriptions, which can be costly and time-consuming. CLIM offers a solution that uses image-text pairs by mosaicking multiple images and treating each image as a \emph{`pseudo region'}. Through this approach, we can obtain an accurate mapping between the pseudo regions and their corresponding text descriptions in the mosaicked image `for free' (at a much lower cost than box labeling).

At each training iteration, we begin by preparing a large canvas that is equally divided into square regions, such as $2\times2$, $3\times3$ or $4\times4$ grids. Next, we sample 4, 9 or 16 image-text pairs from the training dataset, and each image is randomly cropped and then resized to fill a unique square region. In this way, the box location of a pseudo region in the mosaicked image and its corresponding region description are obtained without manual annotations, making it possible to train region-level alignment via contrastive learning.

\subsection{Aligning Region and Text Representations}\label{sec:region_text}
\begin{figure}[t]
    \centering
    \includegraphics[width=1.0\linewidth]{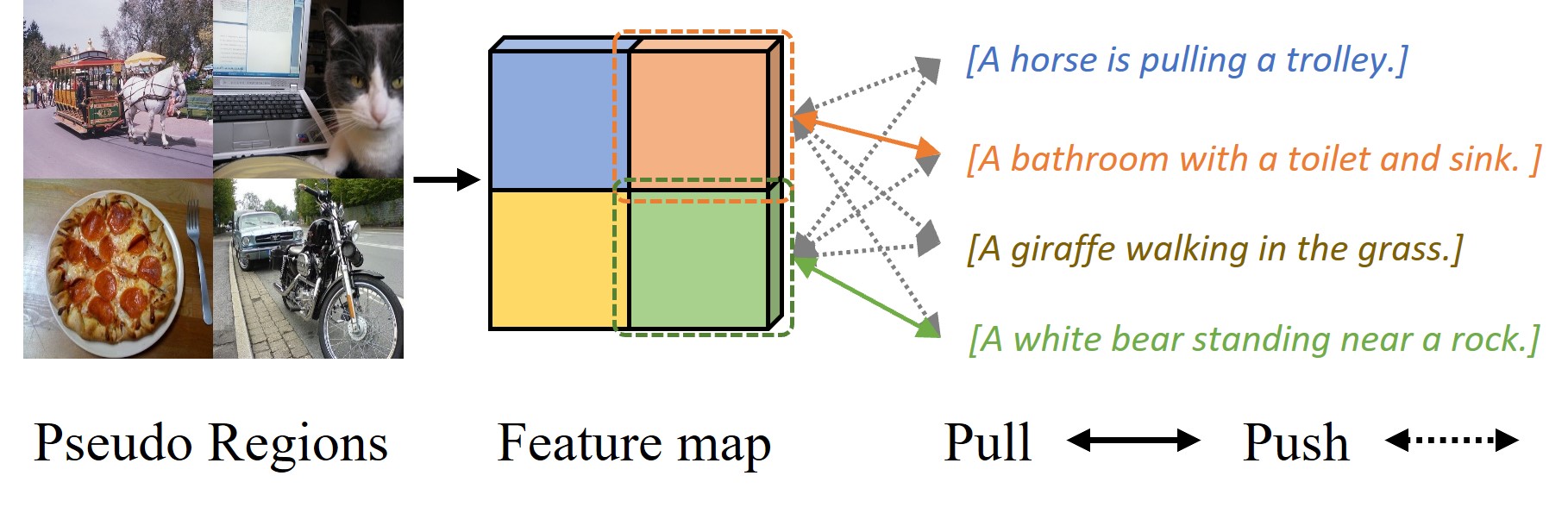}
    \caption{An overview of CLIM. Features of pseudo regions are learned to be similar to the corresponding text embeddings (the colored arrows) and dissimilar to uncorresponding ones (the grey arrows). This process can be applied to both the open-vocabulary object detection (\eg Detic) and the the pre-training of vision-language models (\eg CLIP).}
    \label{fig:method}
\end{figure}

Given a mosaicked image that contains several pseudo regions and their corresponding text descriptions, CLIM aligns the region features with their corresponding text descriptions.
As shown in Figure~\ref{fig:method}, the mosaicked image is first sent to a vision encoder, which outputs the feature map of the mosaicked image.
Then the feature of each pseudo region $f_v$ can be extracted according to the corresponding location.
For the language representation, the text description of each pseudo region is separately fed to the text encoder to obtain the text embedding $f_t$. 

Given region features $f_v$ and their corresponding text embeddings $f_t$, we conduct contrastive learning to force region features to be similar to their text features and dissimilar to those of others. Specifically, the model is trained to maximize the cosine similarity of matched pairs $\langle f_v^{+}, f_t^{+} \rangle$ and minimize the cosine similarity of unmatched pairs $\langle f_v^{+}, f_t^{-}\rangle$.

\subsection{Applications of CLIM}

We apply CLIM to the fine-tuning stage of Detic~\cite{zhou2022detecting} and BARON~\cite{wu2023baron} for open-vocabulary object detection (OVD). Besides, we use CLIM to enhance CLIP~\cite{radford2021learning} model's region representation, and also equip OV-RCNN~\cite{zareian2021open} with CLIM in its vision-language pre-training stage.

\paragraph{OVD with Detic.} Detic~\cite{zhou2022detecting} adopts a two-stage training for open-vocabulary detection. It first trains a detector on base categories, and then fine-tunes the detector using image-level supervision. During the fine-tuning stage, it aligns region feature of the max-size proposal to text embedding of the image label, or aligns the feature of the image box to the text embedding of the caption.
We apply CLIM to Detic by mosaicking images first and aligning the visual features of pseudo regions to the corresponding text embeddings. 
As for the box location for region feature extraction, we follow the original Detic to use the box enclosing the pseudo region for caption loss and use the max-size box inside the pseudo region for image tag (label) loss.

\paragraph{OVD with BARON.} BARON~\cite{wu2023baron} represents an image using a bag of regions. It samples several region proposals in an image and projects region features into word embedding space (pseudo words). Then the pseudo words are concatenated and sent to the text encoder for bag-of-regions embedding.  
The bag-of-regions embeddings are aligned to the caption embeddings of corresponding images. When applying CLIM, we sample region proposals inside a pseudo region (sub-image), and use the bag-of-regions embedding to represent the pseudo region, which is aligned to the corresponding caption embedding using the contrastive loss~\cite{unicl} employed in BARON.

\paragraph{Enhancing CLIP Model's Region Representation.}
Contrastive Language-Image Pretraining (CLIP)~\cite{radford2021learning} learns vision-language alignment on large-scale paired images and texts~\cite{laion5b, cc3m}. For an image-text pair, it obtains a global representation of the image using a [CLS] token in the attention module. CLIM concerns region representations instead of a global representation for the downstream dense prediction applications (\ie object detection). Therefore, we follow the practice of MaskCLIP~\cite{zhou2022maskclip} to extract a feature map from the last attention layer of CLIP's vision model, after which we apply RoIAlign~\cite{he2017mask} on the feature map to extract the region representations. Then we use the same contrastive loss (\ie InfoNCE loss) in CLIP~\cite{radford2021learning} to align region representations and the corresponding text representations. The presence of region-level supervision would significantly enhance the CLIP model's region-text alignment.

\paragraph{Vision-language Pre-training Stage of OV-RCNN.}
OV-RCNN~\cite{zareian2021open} pre-trains a vision backbone and a vision-to-language projection layer that maps region features into word embedding space. It splits an image into $N$ grid regions and calculates the similarity between $N$ grid embeddings and $C$ word embeddings of the caption text. The similarity score of the image-text pair is obtained by averaging the $N \times C$ grid-word cosine similarities, and supervised by a grounding loss (CE loss) that maximizes similarities of matched image-text pairs and minimizes that of unmatched pairs.
When applying CLIM, we divide each pseudo region into grids and average the grid-word similarities to obtain the similarity score between the pseudo region and a text. The same grounding loss is used to impose the alignment between pseudo regions and texts.

\subsection{Discussion}
\paragraph{CLIM v.s. Mosaic Augmentation.} CLIM is reminiscent of the Mosaic augmentation~\cite{bochkovskiy2020yolov4} typically used in conventional object detection tasks, where Mosaic helps manipulate data distribution, increase data diversity and reduce the need for a large batch size. However, CLIM is distinguished from such data augmentation techniques as it is intrinsically a representation learning paradigm that targets at region-language alignment. Data augmentations concern manipulating images in the pre-processing stage, while
CLIM focuses on the contrastive learning between pseudo
regions and corresponding texts during model training to strengthen the region recognition ability. Besides, the data augmentations are specially designed for object detection and require box annotations. On the contrary, CLIM treats images as regions and generally applies to not only open-vocabulary detection methods but also
vision-language pre-training.
\paragraph{Comparison with Manual Labelling.} Compared with CLIM, manually labeling region-text pairs
is quite labor-intensive and inefficient. For example, for 4,000 referring expressions
of regions, the labeling needs 3 weeks of crowdsourcing
as reported in RefCOCO~\cite{kazemzadeh2014referitgame}, which wrapped the labelling process as an online computer game and collected the players' annotations from the web. In contrast, CLIM circumvents such region-level annotation by generating pseudo region-text pairs. 
\begin{table*}[t]
\centering
\caption{Results on open-vocabulary object detection. * means the marked methods are reproduced by us.}
\vspace{-8pt}
\begin{minipage}[t]{0.46\textwidth}
\centering
\subcaption{OV-COCO benchmark}
\vspace{-6pt}
\scalebox{0.88}{\begin{tabular}{l|c|c}
\hline
Method  & Backbone & AP$_{50}^{\mathrm{novel}}$ \\\hline
ViLD~\cite{gu2021open} & RN50 & 27.6\\
OV-DETR~\cite{zang2022open} & RN50 & 29.4\\
RegionCLIP~\cite{zhong2022regionclip} & RN50 &26.8\\
PB-OVD~\cite{gao2021towards} & RN50 &30.8 \\
VLDet~\cite{VLDet}  & RN50 & 32.0\\
F-VLM~\cite{FVLM}  & RN50 &28.0\\
OADP~\cite{wang2023object}  & RN50 & 35.6\\
OV-RCNN~\cite{zareian2021open} &RN50 & 22.8\\
Detic*~\cite{zhou2022detecting} & RN50 & 30.3\\
BARON*~\cite{wu2023baron} & RN50 &34.8\\ \hline
OV-RCNN~\cite{zareian2021open} + CLIM & RN50 & 26.2(+3.4)\\
Detic*~\cite{zhou2022detecting} + CLIM & RN50 & 35.4(+5.1) \\
BARON*~\cite{wu2023baron} + CLIM & RN50& 36.9(+2.1)\\
\hline
\end{tabular}
}
\label{tab:ov_coco_minipage}
\end{minipage}
\hspace{0.04\textwidth}
\begin{minipage}[t]{0.46\textwidth}
\centering
\subcaption{OV-LVIS benchmark}
\vspace{-6pt}
\scalebox{0.88}{
\begin{tabular}{l |c |c}
\hline
Method & Backbone & mAP$_{r}^{\mathrm{mask}}$ \\\hline
ViLD~\cite{gu2021open} &  RN50  & 16.6\\
OV-DETR~\cite{zang2022open} & RN50 & 17.4\\
DetPro~\cite{Du_2022_CVPR} & RN50 &  19.8\\
OC-OVD~\cite{Hanoona2022Bridging} & RN50 &  21.1\\
BARON-KD~\cite{wu2023baron}  & RN50 &22.6 \\
RegionCLIP~\cite{zhong2022regionclip} & RN50  &17.1 \\
OADP~\cite{wang2023object}  & RN50 & 21.7\\
Detic~\cite{zhou2022detecting} & RN50  & 19.5  \\
VLDet~\cite{VLDet} & RN50 & 21.7 \\
F-VLM~\cite{FVLM}*  & RN50x64 &30.1\\ \hline
Detic~\cite{zhou2022detecting} + CLIM & RN50  & 21.8(+2.3) \\
VLDet~\cite{zhou2022detecting} + CLIM & RN50  & 22.2(+0.5) \\
F-VLM~\cite{FVLM}* + CLIM & RN50x64 & 32.3(+2.2) \\

\hline
\end{tabular}}
\label{tab:ov_lvis_minipage}
\end{minipage}
\label{tab:ovd_benchmarks}

\end{table*}

\section{Experiments}\label{sec:experiments}
In the experiment section, we first introduce the main datasets and evaluation metrics. Then we separately introduce the applications to OVD methods (\ie Detic and BARON) and vision-language pre-training (\ie CLIP and OV-RCNN) with their implementation details and ablation study on the design choices. 

\begin{table*}[h]
\centering
\caption{Ablation study of components in CLIM on Detic. Our default setting in the ablation study is marked in blue.}
\begin{minipage}[t]{0.24\textwidth}
\centering
\subcaption{Sanity check 
}
\scalebox{1.0}{  \scalebox{0.85}{
  \begin{tabular}{c|c|c| c}
\hline
\#& Batch Size& CLIM & AP$_{50}^{\mathrm{novel}}$\\
\hline
1 & 4   &  \xmark& 24.4  \\
2 & 20   &\xmark& 27.4 \\
3 & 32   &\xmark&  26.3 \\
\rowcolor{ourscolor}
4 & $\approx$ 20  &\cmark&  32.3 \\

\hline
\end{tabular}
}}
\label{tab:sanity_check}
\end{minipage}
\hspace{0.01\textwidth}
\begin{minipage}[t]{0.24\textwidth}
\centering
\subcaption{Number of pseudo regions}
\scalebox{1.0}{ \scalebox{0.85}{
  \begin{tabular}{c|c| c}
\hline
\#& Grid Size & AP$_{50}^{\mathrm{novel}}$ \\
\hline
1 & $2\times2$ & 30.9 \\
2 & $3\times3$ & 31.2 \\
3 & $4\times4$ & 30.8 \\
\rowcolor{ourscolor}
4 & random     & 32.3 \\
\hline
\end{tabular}
}}
\label{tab:num_subimages}
\end{minipage}
\hspace{0.0\textwidth}
\begin{minipage}[t]{0.24\textwidth}
\centering
\subcaption{Sampling of pseudo regions}
\scalebox{1.0}{ \scalebox{0.95}{
  \begin{tabular}{c|c| c}
\hline
\#& Sampling & AP$_{50}^{\mathrm{novel}}$\\
\hline
1 & text & 31.3	 \\
2 & image & 31.6   \\
\rowcolor{ourscolor}
3 & random & 32.3 \\
\hline
\end{tabular}
}}
\label{tab:region_arrangement}
\end{minipage}
\begin{minipage}[t]{0.24\textwidth}
\centering
\subcaption{Using image tag loss}
\scalebox{1.0}{  \scalebox{0.85}{
  \begin{tabular}{c|c|c| c}
\hline
\# & CLIM & Tag loss & AP$_{50}^{\mathrm{novel}}$\\
\hline
1 &\xmark   & \xmark   &  24.4\\
2 & \xmark  & \cmark &  30.3\\
\rowcolor{ourscolor}
3 & \cmark & \xmark   & 32.3 \\
4 & \cmark & \cmark & 35.4 \\
\hline
\end{tabular}
}}
\label{tab:tag_loss}
\end{minipage}
\label{tab:ablation_detic}
\end{table*}

\subsection{Datasets and Evaluation Metrics} 
We focus on open-vocabulary object detection and report results on the OV-COCO and OV-LVIS benchmarks.
\paragraph{OV-COCO.}
We follow OV-RCNN~\cite{zareian2021open} to divide COCO dataset~\cite{lin2014microsoft} into 48 base classes and 17 novel classes. The training set contains 107,761 images of base category annotations, and the test set contains 4,836 images with both base and novel category annotations. We report the box AP at IoU threshold 0.5, which is denoted as AP$_{50}$. AP$_{50}$ of novel categories (AP$_{50}^{\mathrm{novel}}$) is the major metric to evaluate the OVD performance on OV-COCO benchmark. 

\paragraph{OV-LVIS.}
We follow ViLD~\cite{gu2021open} to use the 866 common and frequent classes in LVIS dataset~\cite{lvis} as base categories, and 337 rare classes as novel categories. Only base category annotations are used during training. For OV-LVIS, we report mean Average Precision (mAP) of masks averaged on IoUs from 0.5 to 0.95. The mAP of rare categories (mAP$_{r}^{\mathrm{mask}}$) is the main evaluation metric for OV-LVIS benchmark.

\subsection{Application to Detic \& BARON}
We apply CLIM to the OVD methods that use caption supervision, \ie Detic~\cite{zhou2022detecting} and BARON~\cite{wu2023baron}. The ablation studies are conducted on Detic using caption loss only on the OV-COCO benchmark as shown in Table~\ref{tab:ablation_detic}.

\paragraph{Implementation Details.}
For Detic~\cite{zhou2022detecting}, we use the Faster RCNN with ResNetC4~\cite{ren2015faster} backbone as the detector on OV-COCO benchmark, and use the detector based on CenterNet2~\cite{zhou2021probablistic} on OV-LVIS benchmark. For the caption supervision, COCO Caption~\cite{cococaption} is used on OV-COCO benchmark and CC3M~\cite{cc3m} is used on OV-LVIS benchmark.
Detic~\cite{zhou2022detecting} alternatively applies image-level supervision and box-level supervision during the fine-tuning stage in its official implementation. We re-implement it by applying the two sources of supervision in parallel and thus requiring only half of the total iterations (45,000). 
We also re-implement BARON~\cite{wu2023baron} in the same way by adopting the two-stage training and parallel supervision. 
Our re-implementation consistently outperforms the results reported in the original papers of Detic~\cite{zhou2022detecting} and BARON~\cite{wu2023baron}.

\paragraph{OV-COCO.}
In Table~\ref{tab:ov_coco_minipage}, we report the results of applying CLIM to Detic and BARON on OV-COCO benchmark. When applied to the image-level supervision of Detic and BARON at the fine-tuning stage, CLIM increases the performance on novel categories by $5.1$ AP$_{50}$ and $2.1$ AP$_{50}$, respectively.

\paragraph{OV-LVIS.}
On OV-LVIS benchmark, we apply CLIM to Detic~\cite{zhou2022detecting} and its follow-up VLDet~\cite{VLDet}. As shown in Table~\ref{tab:ov_lvis_minipage}, CLIM improves Detic by 2.3 mAP$_{r}^{\mathrm{mask}}$. CLIM also improves VLDet that is built upon Detic and matches region proposals with object concepts to produce region-text pairs. This indicates that CLIM is orthogonal to methods that adopt pseudo-labelling strategies to obtain object-language correspondence.

\paragraph{Sanity Check.} We verify that the improvement of CLIM is not mainly attributed to the increased number of images. 
  In our implementation of Detic, we set the batch size of box and caption supervision as 2 and 4 on each GPU, respectively. When applying CLIM to Detic, we further split the 4 images for caption supervision into 2 with mosaic and 2 without mosaic. And we randomly choose $2\times2$, $3\times3$ and $4\times4$ mosaic. Therefore, there are on average $2 + 2\times(4+9+16) / 3 \approx 20$ images for caption supervision in each iteration. 
  As the sanity check, we replace CLIM with simply stacking more images for caption supervision in a batch. As shown in Table~\ref{tab:sanity_check}, increasing the batch size of caption supervision from 4 to 20 only leads to 3.0 performance gain on novel categories, while CLIM achieves 7.9 performance gain (32.3 AP$_{50}^{\mathrm{novel}}$ v.s. 24.4 AP$_{50}^{\mathrm{novel}}$) under this fair comparison. Besides, we also observe that further increasing the batch size does not bring any improvement in Table~\ref{tab:sanity_check} (\#4).

\paragraph{Number of Pseudo Regions.} In Table~\ref{tab:num_subimages}, we study the number of pseudo regions in a mosaicked image. The overall resolution of the mosaicked image is fixed as $800 \times 800$ in this ablation study. We first separately choose $2 \times 2$, $3 \times 3$ and $4 \times 4$ and observe that the $3 \times 3$ mosaic achieves the best performance on novel categories (31.2 AP$_{50}^{\mathrm{novel}}$). However, the performance gap between these single-pattern choices is marginal. When we randomly choose from the three settings in each iteration, the AP$_{50}$ on novel categories grows by 1.1. 
This indicates that the mixed use of $2 \times 2$, $3 \times 3$ and $4 \times 4$ mosaic allows
the model to generalize to different region patterns, thus improving the performance on novel categories.
However, simply increasing the number of pseudo regions does not provide consistent performance gain (\#2 and \#3).

\paragraph{Sampling of Pseudo Regions.} By default, we randomly combine pseudo regions. And we also consider grouping samples that are similar in CLIP text representations or image representations by calculating cosine distance. However, both approaches decrease the performance as shown in Table~\ref{tab:region_arrangement}. Combining pseudo regions that have similar text descriptions or visual contents tends to generate mosaciked images
of a specific pattern, limiting the models’ generalization ability to different region patterns in testing.

\paragraph{Image Tag Loss in Detic.}
Detic can be trained with caption loss only, or with both max-size image tag (label) loss and the caption loss. We report the results before and after adding image tag loss in Table~\ref{tab:tag_loss}. We show that the image tag loss improves the baseline to 30.3 AP$_{50}^{\mathrm{novel}}$ (\#1 and \#2), and also improves the performance of our CLIM to 35.4 AP$_{50}^{\mathrm{novel}}$ (\#3 and \#4). This also indicates that CLIM can bring consistent performance gain on different variants of Detic.

\subsection{Enhancing CLIP's Region Representation}
We study the application of CLIM to CLIP models~\cite{radford2021learning} and analyze how CLIM would help improve CLIP's region representation. For simplicity, we use the Top-1 and Top-5 accuracy of classifying the COCO dataset's bounding boxes to evaluate the enhancement of region representation. In addition to the zero-shot region classification, we also build open-vocabulary detectors upon the models trained by CLIM following F-VLM~\cite{FVLM} to further validate the enhancement of region representation for realistic downstream application.
\paragraph{Implementation Details.}
When applying CLIM to CLIP models, we mainly study the ViT-B-16 variant of CLIP and use the model weights released by OpenAI to initialize our training. For the experiment on OV-COCO, we train the CLIP model on COCO Caption~\cite{cococaption} for 100 epochs. For the experiment on OV-LVIS, we train the CLIP model on CC3M~\cite{cc3m} for 3 epochs. Following the pre-training of the original CLIP models~\cite{radford2021learning}, we use AdamW optimizer and set the batch size to 128 and the learning rate to 1e-5. After being trained by CLIM, the CLIP models are then used to build open-vocabulary detectors.

\paragraph{Image Resolution for CLIM.} We experiment with different resolutions ($320 \times 320$, $640 \times 640$, $1024 \times 1024$) of the mosaicked images to train the model. As shown in Table~\ref{tab:image_resolution}, CLIM significantly improves the region representation of CLIP, and increasing input resolution consistently produces performance gains.
However, we do not further enlarge the input size and apply $1024 \times 1024$ as image resolution for training due to the quartically increasing computation cost. 
\begin{table}[t]
  \small
  \centering
\caption{Resolution of the mosaicked images.}
  {  \scalebox{1.05}{
  \begin{tabular}{c|c|c| cc}
\hline
\#& Data & Resolution & Top1 & Top5 \\
\hline
1 & COCO Caption&-   & 29.2  & 51.6  \\
2 &COCO Caption &$320 \times 320$  & 57.8  & 80.0  \\
3 &COCO Caption &$640 \times 640$  & 61.3 & 83.8 \\
\rowcolor{ourscolor}
4 & COCO Caption&$1024 \times 1024$  &  62.2 & 84.3 \\
\hline
\end{tabular}
}}
\label{tab:image_resolution}
\end{table}

\paragraph{Building Open-Vocabulary Object Detector.} In addition to zero-shot inference on classifying ground truth bounding boxes, we also built open-vocabulary detectors following the architecture of F-VLM~\cite{FVLM}, to verify the enhancement of region representation. As shown in Table~\ref{tab:ov_lvis_minipage}, with the RN50x64 model trained by CLIM, we improve F-VLM by 2.2 mAP$^{_\mathrm{mask}}_r$ on the OV-LVIS benchmark. Besides, we also build F-VLM with the ViT-B-16 model on the OV-COCO and OV-LVIS benchmarks as shown in Table~\ref{tab:ovd_clip}. To obtain multi-scale feature maps for the Feature Pyramid Network (FPN) in the ViT-based detector, we extract feature maps from the 3th, 5th, 7th and 11th attention layers of the ViT model, and interpolate them to $[\frac{1}{4}, \frac{1}{8}, \frac{1}{16}, \frac{1}{32}]$ of the input image size. 
We observe weak performances of the ViT-based detector on both benchmarks for both base and novel categories when using the original ViT-B-16 model released by OpenAI, which are significantly improved by the model trained by CLIM (\#2). 
Due to the lack of translation invariance and equivariance in the transformer architecture, the presence of region-level supervision by CLIM is particularly necessary during the pre-training of ViT-based vision-language models for downstream dense prediction tasks like object detection.

\begin{table}[t]
  \small
  \centering
\caption{Open-vocabulary detection results of applying CLIM to CLIP's ViT-B-16 model.}
  {  \scalebox{0.92}{
\begin{tabular}{l|cc|ccc}
  \hline
  \multirow{2}{*}{Model}&\multicolumn{2}{c|}{OV-COCO}&\multicolumn{3}{c}{OV-LVIS}\\
&AP$_{50}^{\mathrm{novel}}$ & AP$_{50}^{\mathrm{base}}$& mAP$_{r}^{\mathrm{mask}}$&mAP$_{c}^{\mathrm{mask}}$ &mAP$_{f}^{\mathrm{mask}}$ \\ \hline
 CLIP &21.6&36.4&14.8&20.5&26.1 \\ 
 \rowcolor{ourscolor}
 CLIM &\textbf{25.7}& \textbf{42.5}& \textbf{20.8}&\textbf{25.6}&\textbf{29.7} \\
\hline
\end{tabular}
}}
\label{tab:ovd_clip}
\end{table}

\begin{table}[t]
  \small
  \centering
\caption{Comparison with RegionCLIP on zero-shot region classification and open-vocabulary object detection. 
RegionCLIP applies the well-trained RPN and pre-defined vocabulary of object nouns for region-text alignment.}
  {  \scalebox{0.95}{
\begin{tabular}{l|c|cc|cc}
  \hline
  \multirow{2}{*}{\#} &  \multirow{2}{*}{Method}&\multicolumn{2}{c|}{Region Classification}&\multicolumn{2}{c}{OV-COCO}\\
  &&Top1&Top5& AP$_{50}^{\mathrm{novel}}$ & AP$_{50}^{\mathrm{base}}$\\ \hline
1& CLIP &29.2&51.6&21.6&36.4 \\
2& RegionCLIP &62.8&84.7& 26.1 & 42.4 \\ 
\rowcolor{ourscolor}
3 & CLIM &62.2&84.3& 25.7 &42.5 \\
\hline
\end{tabular}
}}
\label{tab:regionclip}
\end{table}

\paragraph{Comparison with RegionCLIP.} We compare our approach with RegionCLIP~\cite{zhong2022regionclip} in the ability of enhancing region representation in Table~\ref{tab:regionclip}. RegionCLIP matches region proposals with object nouns to generate the region-text pairs for the learning of region-language alignment. 
We implement RegionCLIP using ViT-B-16 and COCO Caption dataset~\cite{cococaption}. The region proposals are detected by an region proposal network (RPN) trained on COCO's box annotations of base categories, which are cropped and sent to CLIP's image encoder (ViT-B-16) to obtain region representations. The cosine similarities between region representations and text representations of object nouns are used as the metric for matching. We also set the input image resolution as $1024 \times1024$ to train RegionCLIP. It is noticeable that RegionCLIP is built on a strong assumption of the existence of a well-trained RPN and a pre-defined vocabulary of object nouns~\cite{zhong2022regionclip}. However, we still achieve comparable results on both zero-shot region classification and open-vocabulary object detection.

\subsection{Application to OV-RCNN}
For OV-RCNN~\cite{zareian2021open}, we follow the official implementation, which pre-trains the model for 40,000 iterations with a batch size of 64 on COCO Caption~\cite{cococaption} and finetunes it for 150,000 iterations with a batch size of 8 on the box annotations of base categories in COCO dataset~\cite{lin2014microsoft}. Our CLIM is applied in the pre-training stage.
As shown in Table~\ref{tab:ov_coco_minipage}, CLIM boosts the final performance of OV-RCNN on novel categories by 3.4 AP$_{50}$. 


\section{Visualization \& Analysis}\label{sec:analysis}

We provide visualization and analysis of the enhancement of region representation in this section. First, we analyze the open-vocabulary object detector (Detic) trained with CLIM. Then we visualize how CLIM improves the vision-language alignment of CLIP's region representation.

\begin{figure}[t]
    \centering
    \includegraphics[width=1.0\linewidth]{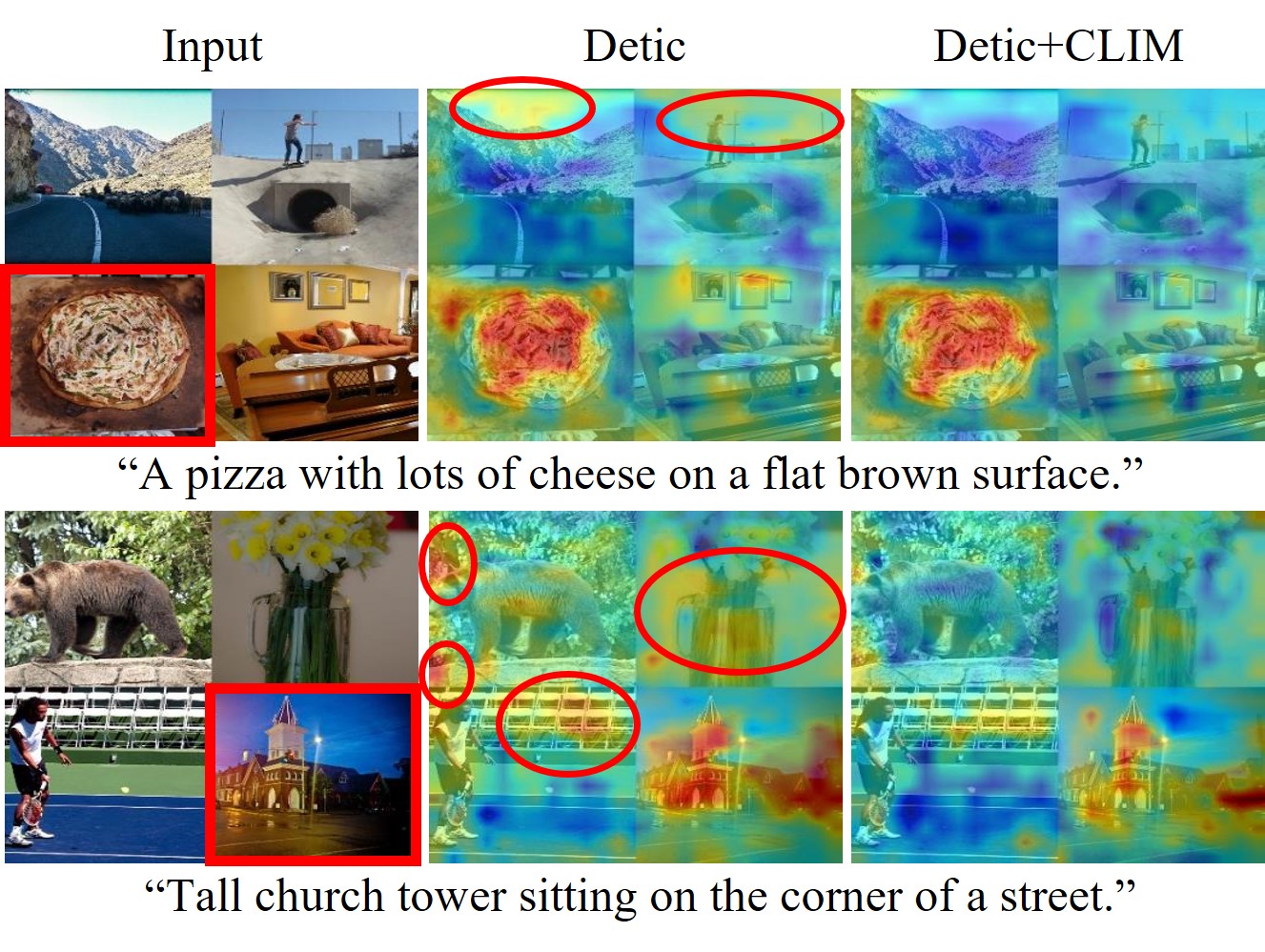}
    \caption{Feature map response on mosaicked images. The sub-images with a red border correspond to the text descriptions below. False positive regions with high response values are highlighted with red circles. }
    \label{fig:mosaic_images}
\end{figure}

\begin{figure}[t]
    \centering
\includegraphics[width=1.0\linewidth]{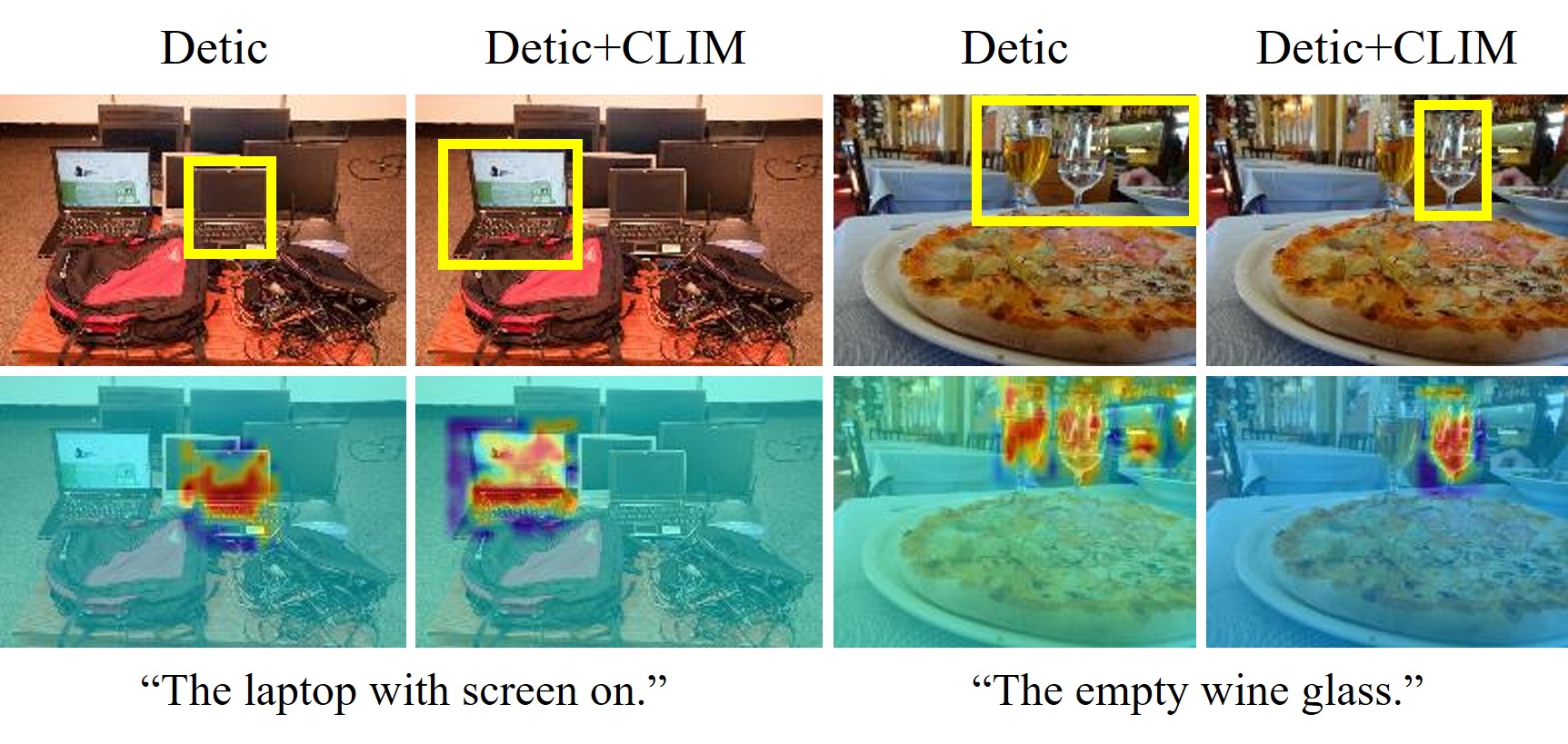}
    \caption{Feature map response on natural images. The yellow boxes are the detected bounding boxes of the queried text descriptions. }
    \label{fig:natural_images}
\end{figure}

\subsection{Region Response to Text Description}
As images are taken as pseudo regions during the training of CLIM, the models should have earned the generalization ability to localize regions given corresponding text descriptions. 
Specifically, we analyze the models' ability to response to queried texts at the regions of interest by calculating cosine similarities between feature map and the text embeddings. We compare the features map response of the Detic baseline and the Detic trained with CLIM. And we consider two types of images, \ie mosaicked images that only appear during training and natural images that are ubiquitous in testing.

\paragraph{Mosaicked Images.} As shown in Figure~\ref{fig:mosaic_images}, the feature map response of our model is more concentrated on the pseudo region corresponding to the queried text. In comparison, the distribution of the Detic baseline's feature map response is more diffused. And there are many high responses outside of the queried pseudo regions as highlighted by red circles in Figure~\ref{fig:mosaic_images}.

\paragraph{Natural Images.} For the natural images, we not only visualize the feature map response but also show if the detector can localize the queried text by bounding boxes. This is similar to the task of referring expression comprehension~\cite{yu2016modeling}, where  the model is required to accurately locate the desired object given a text description that contains the attributes or the context of the object. 
As shown in Figure~\ref{fig:natural_images}, the Detic baseline model detects the undesired objects or cannot accurately locate the objects. Although the Detic baseline model has learned the vision-language alignment at the image level, it cannot effectively transfer the alignment knowledge to regions. 
In comparison, our CLIM, which forms pseudo regions by mosaicking images, forces the detector to learn to associate regions with texts.

\begin{figure}[t]
    \centering
    \includegraphics[width=1.0\linewidth]{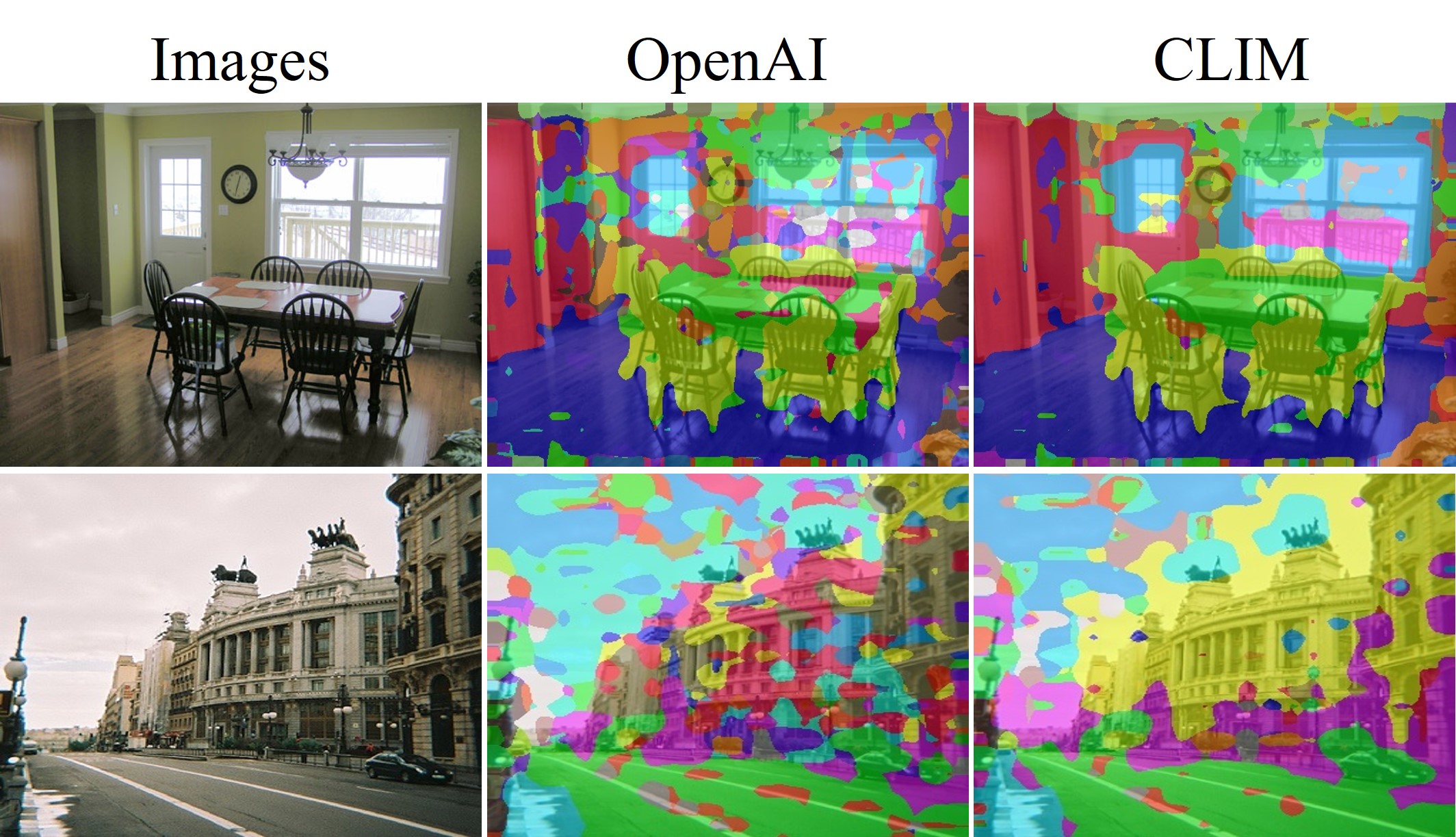}
    \caption{Visualization of CLIP's feature map by per-pixel classification. The images are from PASCAL Context~\cite{everingham2010pascal} dataset with 59 pixel categories.}
    \label{fig:maskclip}
\end{figure}

\subsection{Enhancement of CLIP's Region Representation}
To verify CLIM's effectiveness in improving CLIP's region representation, we visualize the vision-language alignment of CLIP's feature map using per-pixel classification. Specifically, we choose the ViT-B-16 and compare the original model released by OpenAI with the model trained with CLIM. The images are from PASCAL Context dataset~\cite{everingham2010pascal} with 59 pixel categories.
As shown in Figure~\ref{fig:maskclip}, the feature map visualization of CLIM model is less diffused and different objects are more accurately divided, indicating the significant improvement of vision-language alignment on CLIP model's feature map. CLIM benefits per-pixel recognition even though the single pixel embeddings are not directly supervised during training.

\section{Conclusion}\label{sec:conclusion}

In this paper, we present a novel method, Contrastive Language-Image Mosaic (CLIM), to exploit large-scale image-text pairs for region-language alignment without needing expensive bounding box annotations or relying on inaccurate box predictions. CLIM achieves this by mosaicking multiple images and treating each image as a \emph{`pseudo region'} within the context of the mosaicked image, and then learning region-level representations via contrastive learning.
By eliminating the need for costly annotations or noisy box predictions, CLIM presents an efficient and general solution with only image-text pairs for training open-vocabulary object detectors as well as improving vision-language model's region representation.

\setcounter{table}{0}
    \renewcommand{\thetable}{A\arabic{table}}
\setcounter{figure}{0}
\renewcommand{\thefigure}{A\arabic{figure}}
\setcounter{section}{0}
\renewcommand{\thesection}{A\arabic{section}}

\section{Appendix}

\subsection{Re-implementations}

\paragraph{Detic \& BARON.}
Detic~\cite{zhou2022detecting} alternatively applies image-level supervision and box-level supervision during the fine-tuning stage in its official implementation. We re-implement it upon MMDetection~\cite{mmdetection} and apply the two sources of supervision in parallel, thus requiring only half of the total iterations (45,000). 
We also re-implement BARON~\cite{wu2023baron} in the same way by adopting the two-stage training and parallel supervision. As shown in Table~\ref{tab:re_implement}, our re-implementation improves Detic and BARON by 2.5 and 1.7 AP$_{50}^{\mathrm{novel}}$, respectively.
\paragraph{F-VLM.} We also re-implement F-VLM~\cite{FVLM} upon MMDetection~\cite{mmdetection} following the setting of Table 3 (2.88k iterations) in the F-VLM's paper. As shown in Table~\ref{tab:reimplement_fvlm}, our re-implementation (30.1 mAP$_{r}^{\mathrm{mask}}$) outperforms the reported result (27.7 mAP$_{r}^{\mathrm{mask}}$) by a large margin.

\subsection{Application Details}

In this section, we provide more details on the applications of CLIM to Detic~\cite{zhou2022detecting}, BARON~\cite{wu2023baron}, CLIP~\cite{radford2021learning} and OV-RCNN~\cite{zareian2021open}.
\subsubsection{Application to Detic}
There are two variants of Detic that use caption loss and image tag loss, respectively.
\paragraph{Caption Loss.} Detic originally uses the image box for caption loss, and aligns the feature of the image box to the text embedding of the corresponding caption. Similarly, we obtain the box that tightly encloses the pseudo region for caption loss when applying CLIM to Detic as shown in Figure~\ref{fig:detic_clim}(a).

\begin{table}[h]
  \small
  \centering
\caption{Re-implementation of Detic~\cite{zhou2022detecting} and BARON~\cite{wu2023baron} on OV-COCO. Our re-implementation achieves higher baseline performance than those reported in the papers.}
  {  \scalebox{0.83}{
  \begin{tabular}{c|c|c| c >{\color{gray}}c >{\color{gray}}c}
\hline
Method & Re-implement & Tag Loss & mAP$_{50}^{\mathrm{novel}}$ & mAP$_{50}^{\mathrm{base}}$ & mAP$_{50}$\\
\hline
Detic & \xmark & \xmark   &  21.0 & 51.9  &43.8\\
\rowcolor{ourscolor}
Detic & \cmark  & \xmark   &  24.4 & 54.6& 46.7\\
Detic & \xmark   & \cmark &27.8&  51.1 &45.0 \\
\rowcolor{ourscolor}
Detic & \cmark & \cmark   & 30.3 & 54.7 & 48.3\\
BARON & \xmark & \xmark &33.1&54.8 &49.1\\
\rowcolor{ourscolor}
BARON & \cmark & \xmark & 34.8 & 55.0 &49.7\\
\hline
\end{tabular}
}}
\label{tab:re_implement}
\end{table}

\begin{table}[h]
  \small
  \centering
\caption{Re-implementation of F-VLM~\cite{FVLM} on OV-LVIS. Our re-implementation achieves higher baseline performance than reported result in the paper.}
  {  \scalebox{0.95}{
\begin{tabular}{l|c| c>{\color{gray}}c >{\color{gray}}c}
\hline
Method & Re-implement & mAP$_{r}^{\mathrm{mask}}$ & mAP$_{c}^{\mathrm{mask}}$&mAP$_{f}^{\mathrm{mask}}$\\\hline
F-VLM & \xmark &27.7&-&- \\
\rowcolor{ourscolor}
F-VLM & \cmark &30.1&25.9&27.7\\
\hline
\end{tabular}}}
\label{tab:reimplement_fvlm}
\end{table}

\paragraph{Image Tag Loss.}
 Detic originally uses the max-size box for image tag (label) loss, and aligns the feature of the max-size box to the text embeddings of corresponding image tags (labels). We obtain the max-size box inside each pseudo region for the image tag loss when applying CLIM to Detic as shown in Figure~\ref{fig:detic_clim}(b). As the image tags may contain multiple object categores, the feature of a pseudo region can be aligned to multiple text embeddings.

\subsubsection{Application to BARON}

\begin{figure*}[t!]
\begin{minipage}[t]{1.0\textwidth}
    \centering
    \includegraphics[width=1.0\linewidth]{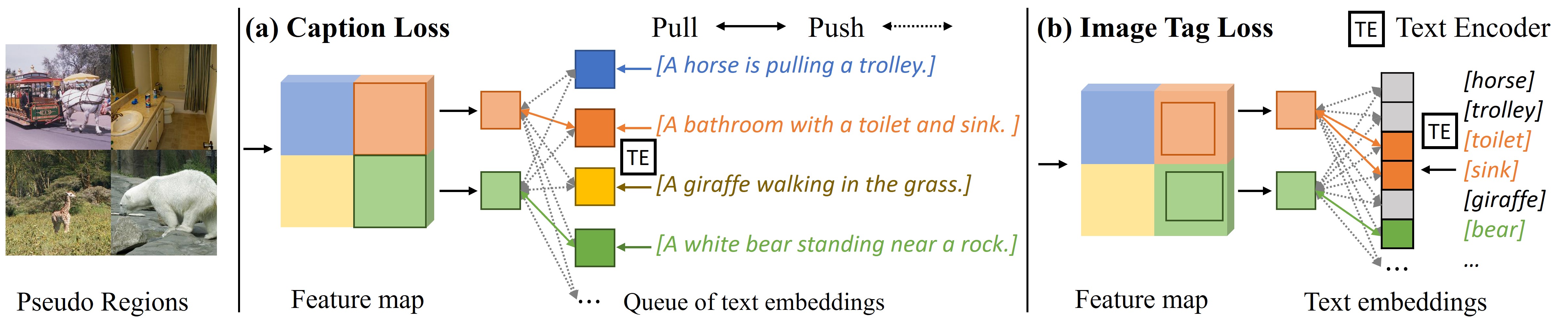}
    \caption{Application of CLIM to Detic. The feature of a pseudo region is aligned to the corresponding text embedding. As the image tags may contain multiple object categories, the feature of a pseudo region can be aligned to multiple text embeddings at the same time as shown in \textbf{(b)}.}
    \label{fig:detic_clim}
\end{minipage}
\begin{minipage}[t]{1.0\textwidth}
    \centering
    \includegraphics[width=1.0\linewidth]{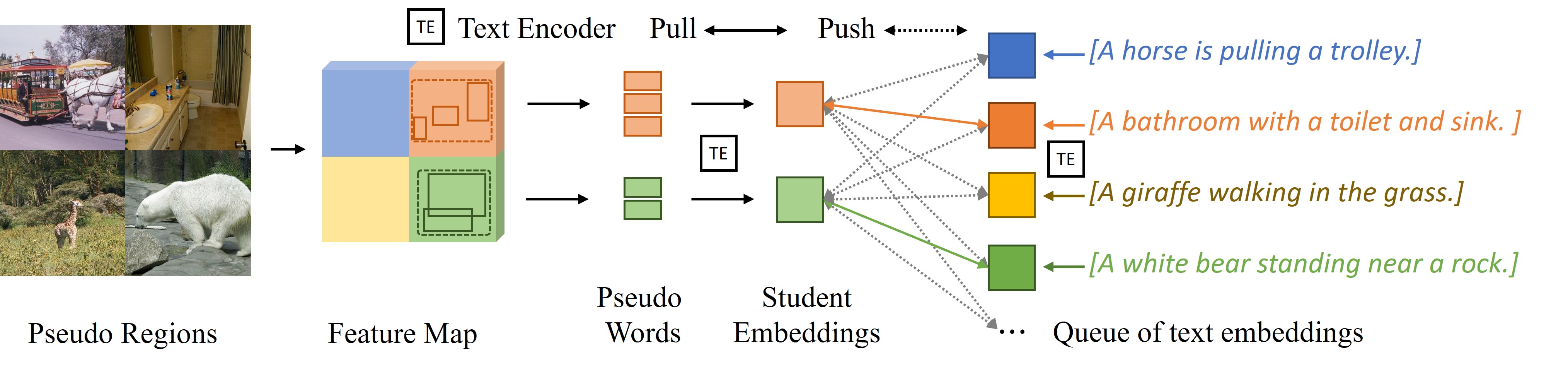}
    \caption{Application of CLIM to BARON. A dashed box stands for a bag of regions. 
    The bag-of-regions embedding (student embedding) of a pseudo region is aligned to the corresponding text embedding.
    }
    \label{fig:baron_clim}
\end{minipage}
\begin{minipage}[t]{1.0\textwidth}
    \centering
    \includegraphics[width=1.0\linewidth]{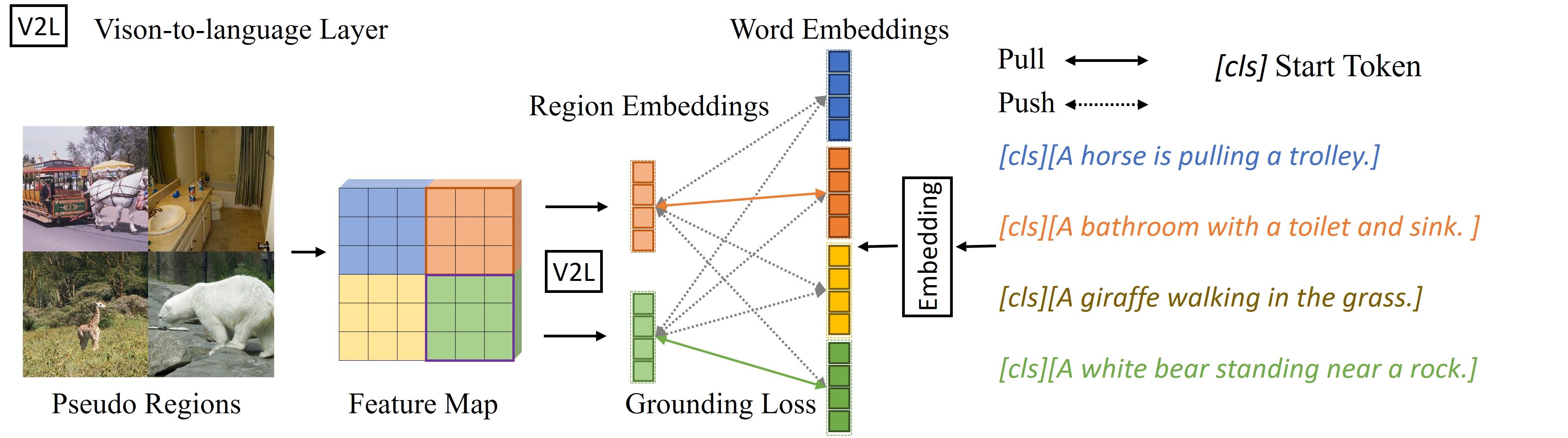}
    \caption{Application of CLIM to OVR-CNN. The similarity of a pseudo region-text pair is the grounding score between the grid region embeddings and word embeddings.}
    \label{fig:ovrcnn_clim}
\end{minipage}
\end{figure*}

BARON samples a bag of regions to represent the content of an image. It projects region features to word embedding space (pseudo words). As shown in Figure~\ref{fig:baron_clim}, it concatenates the pseudo words and sends them to the text encoder to obtain the bag-of-regions embeddings (student embeddings).
When applying CLIM to BARON, we sample a bag of regions inside a pseudo region (sub-image) to represent the content of the pseudo region.

\subsubsection{Application to CLIP}
Given the simplicity of the CLIP method~\cite{radford2021learning}, we do not additionally give a figure for training as it would be the same as Figure 2 in the main paper. Instead, we provide the details for region feature extraction. We first briefly introduce the image-level representation of CLIP and then depict how it is adapted for region representation.

\paragraph{CLIP's Image Representation.} The image representation is obtained from the last residual attention block of CLIP. 
The input to the last residual attention block is $\bm{x} = (x_0, x_1, ..., x_{h \times w})^\mathsf{T}$ representing $h\times w$ image patches $\{x_{i}| i\in \{1, 2, ..., h\times w\}\}$ and a class embedding $x_0$. A residual attention block $\bm{z}=\text{ResAttn}(\bm{x})$ can be written as:
\begin{align*}
    \bm{q} =& \text{Emb}_q(\bm{x}), k =\text{Emb}_k(\bm{x}), v = \text{Emb}_v(\bm{x}) \\
    \bm{y} = &\bm{x} + \text{Proj}(\text{SoftMax}(\frac{\bm{qk}}{c})\cdot \bm{v}) \\
    \bm{z} =& \bm{y} + \text{FFN}(\bm{y}),
\end{align*}
where $c$ is a constant, Proj represents a projection layer, Emb comprises a layer norm and a projection layer, and FFN stands for a feed forward network. Therefore the image representation of the whole input is $\bm{x}_{\text{image}} = \text{Emb}_{\text{out}}(\bm{z})[0]$.

\paragraph{CLIP's Region Representation.} To extract region representation, we first obtain a feature map (dense representation) of the whole image following MaskCLIP~\cite{zhou2022maskclip}, which slightly modifies the last residual attention block. Specifically, it keeps all the projection layers, layer norms and FFNs while discarding the self-attention. The modified residual attention block $\bm{z}'=\text{ModifiedResAttn}(\bm{x})$ can be written as:
\begin{align*}
     \bm{v} =& \text{Emb}_v(\bm{x}) \\
    \bm{y}' = &\bm{x} + \text{Proj}(\bm{v}) \\
    \bm{z}' =& \bm{y}' + \text{FFN}(\bm{y}').
\end{align*}

Therefore, we can obtain an $h \times w$ feature map as $\mathcal{X}_{\text{dense}} = \text{Reshape}(\text{Emb}_{\text{out}}(\bm{z}')[1\colon h \times w])$. Given the bounding box of a region, we can obtain the region representation by 
applying a $1 \times 1$ RoIAlign on the feature map $\mathcal{X}_{\text{dense}}$.

\begin{figure*}[h]
    \centering
    \includegraphics[width=1.0\linewidth]{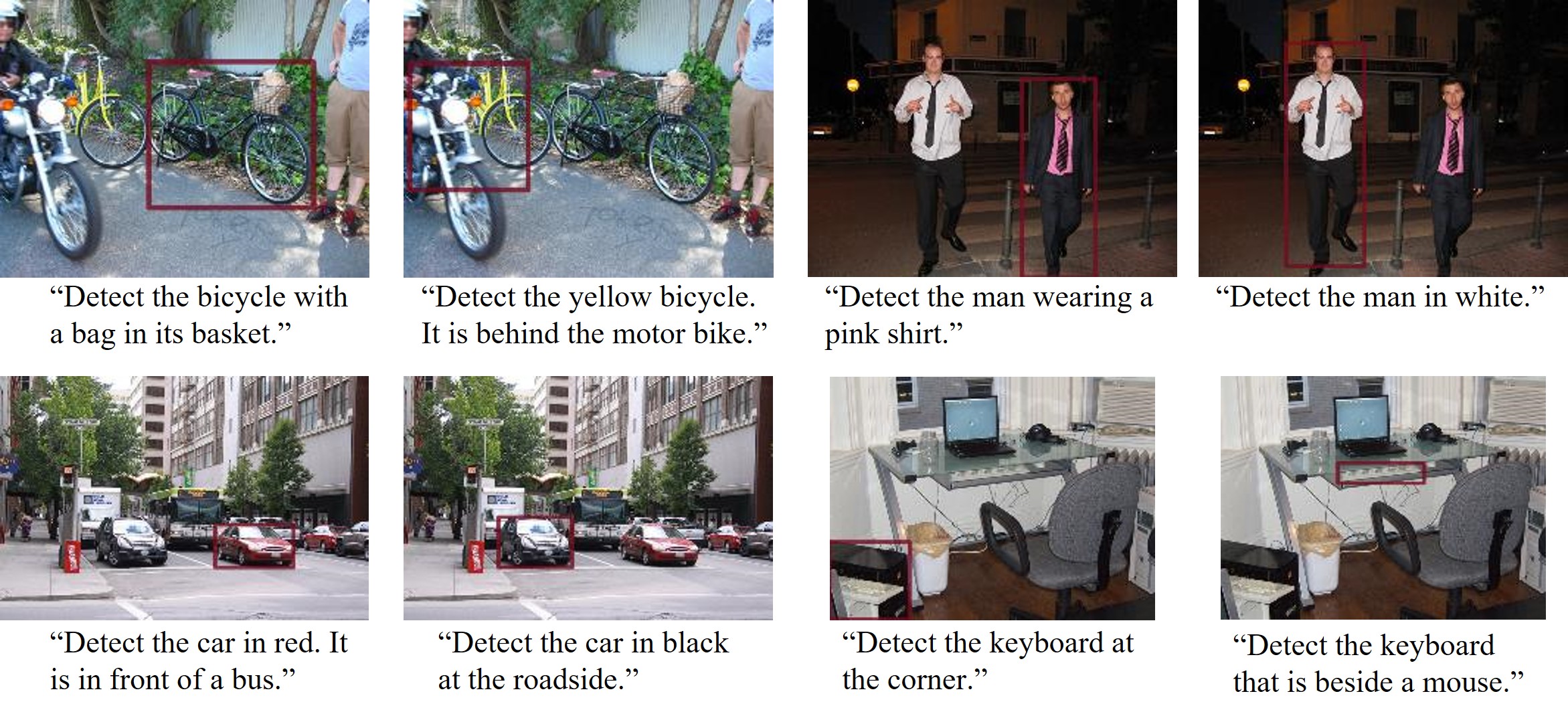}
    \caption{Localization of text queries. The texts below the images are the descriptions of the queried objects.
    }
    \label{fig:more_vg_results}
\end{figure*}

\subsubsection{Application to OVR-CNN} 
In the pre-training stage of OVR-CNN~\cite{zareian2021open}, the feature map is split into a grid of $N \times N$ regions and $S$ regions are sampled. Empirically $S$ is set as 100. The similarity score between an image-text pair is measured by a grounding score between the $S$ region embeddings and word embeddings of the image caption as shown in Figure~\ref{fig:ovrcnn_clim}. When applying CLIM to OVR-CNN, we sample $S$ regions inside each pseudo region to calculate the region-text similarity using the grounding score. The $S$ region embeddings and the word embeddings of all the corresponding captions are used to calculate the grounding score. 
Besides, there are an image-text matching (ITM) loss and a masked language modelling (MLM) loss defined in OVR-CNN. The two losses remain the same in our implementation. For the calculation of grouding score and the detailed definition of the ITM and MLM losses, please refer to the paper of OVR-CNN~\cite{zareian2021open}.

\subsection{Composition of Pseudo Regions}
In addition to the single pseudo regions, we can also combine neighboring pseudo regions into a new pseudo region whose contents are more diverse and complicated. We name this type of pseudo region as `composed region'. Inside a mosaicked image, we
randomly sample G groups of spatially continuous pseudo regions as the composed regions, where G = 4 in practice. The feature of a composed region is aligned to texts of all the constituent sub-images simultaneously, which requires a multi-label loss.
Composed pseudo regions are applied to Detic~\cite{zhou2022detecting} and BARON~\cite{wu2023baron} as these methods originally use multi-label losses, \ie BCE loss in Detic and Soft Target CE loss~\cite{unicl} in BARON. As shown in Table~\ref{tab:composition}, the use of composed regions brings 2.1 AP$_{50}^{\mathrm{novel}}$ performance gain on Detic.

\begin{table}[h]
  \small
  \centering
\caption{The effect of composed regions.}
  {  \scalebox{1.0}{
  \begin{tabular}{c|c|c| c>{\color{gray}}c}
\hline
\#& Method&Composition & AP$_{50}^{\mathrm{novel}}$ & AP$_{50}^{\mathrm{base}}$ \\
\hline
 1 &Detic &\xmark  & 30.2	& 55.1   \\
 \rowcolor{ourscolor}
3 & Detic&\cmark & 32.3	& 54.1    \\

\hline
\end{tabular}
}}
\label{tab:composition}
\end{table}

\subsection{Localization of Text Queries}
In Figure~\ref{fig:more_vg_results}, we show more examples of using our model (Detic+CLIM) trained on COCO dataset for localizing queried text descriptions. The descriptions of the queried objects are finer-grained compared to the open-vocabulary detection inference.

 \subsection{Detection Results}
\begin{figure*}[t!]
\begin{minipage}[t]{1.0\textwidth}
    \centering
    \includegraphics[width=1.0\linewidth]{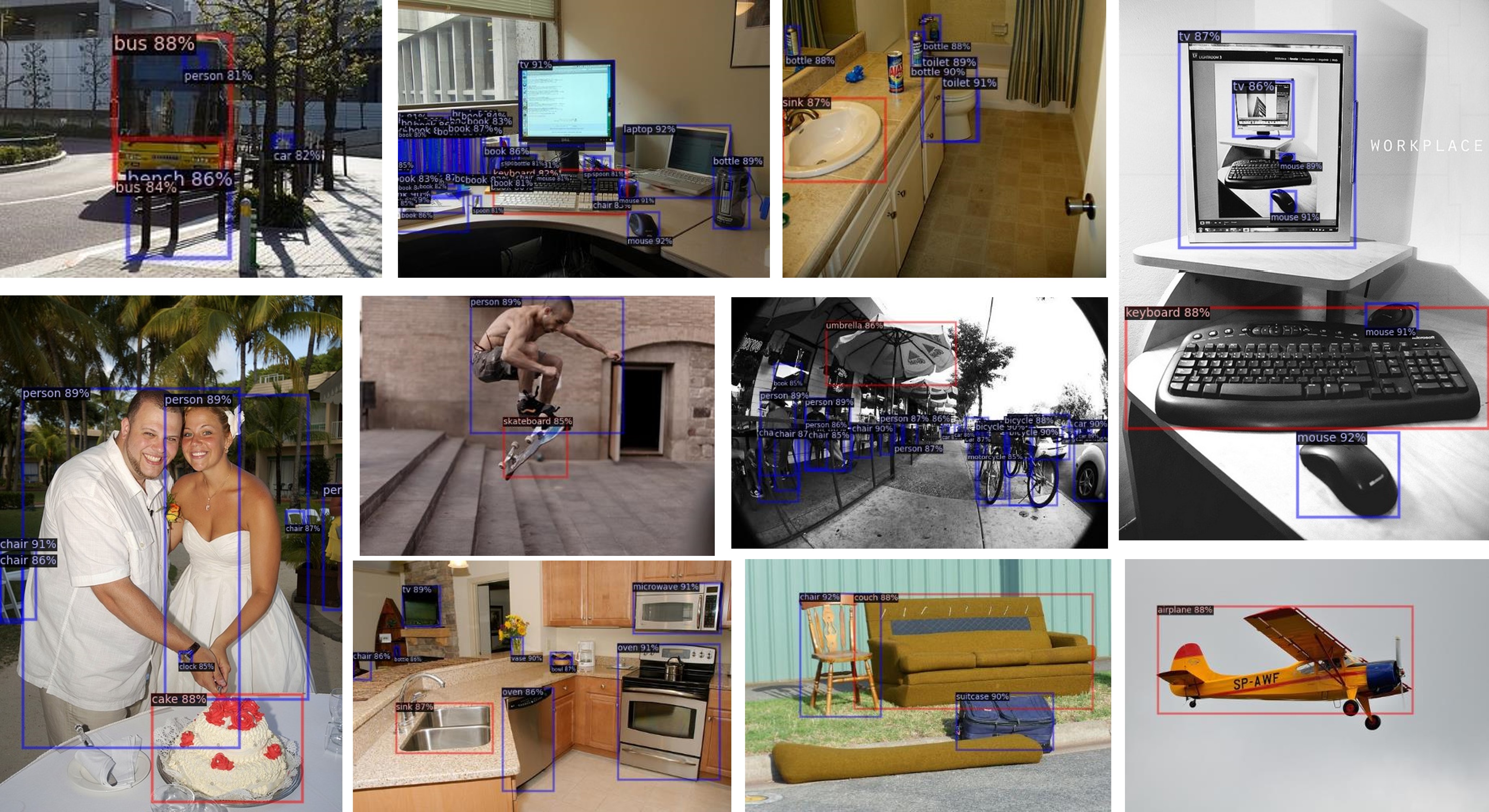}
    \caption{Visualization of detection results on OV-COCO. Red boxes are for novel categories.
    }
    \label{fig:coco_dets}
\end{minipage}
\begin{minipage}[t]{1.0\textwidth}
        \centering
    \includegraphics[width=1.0\linewidth]{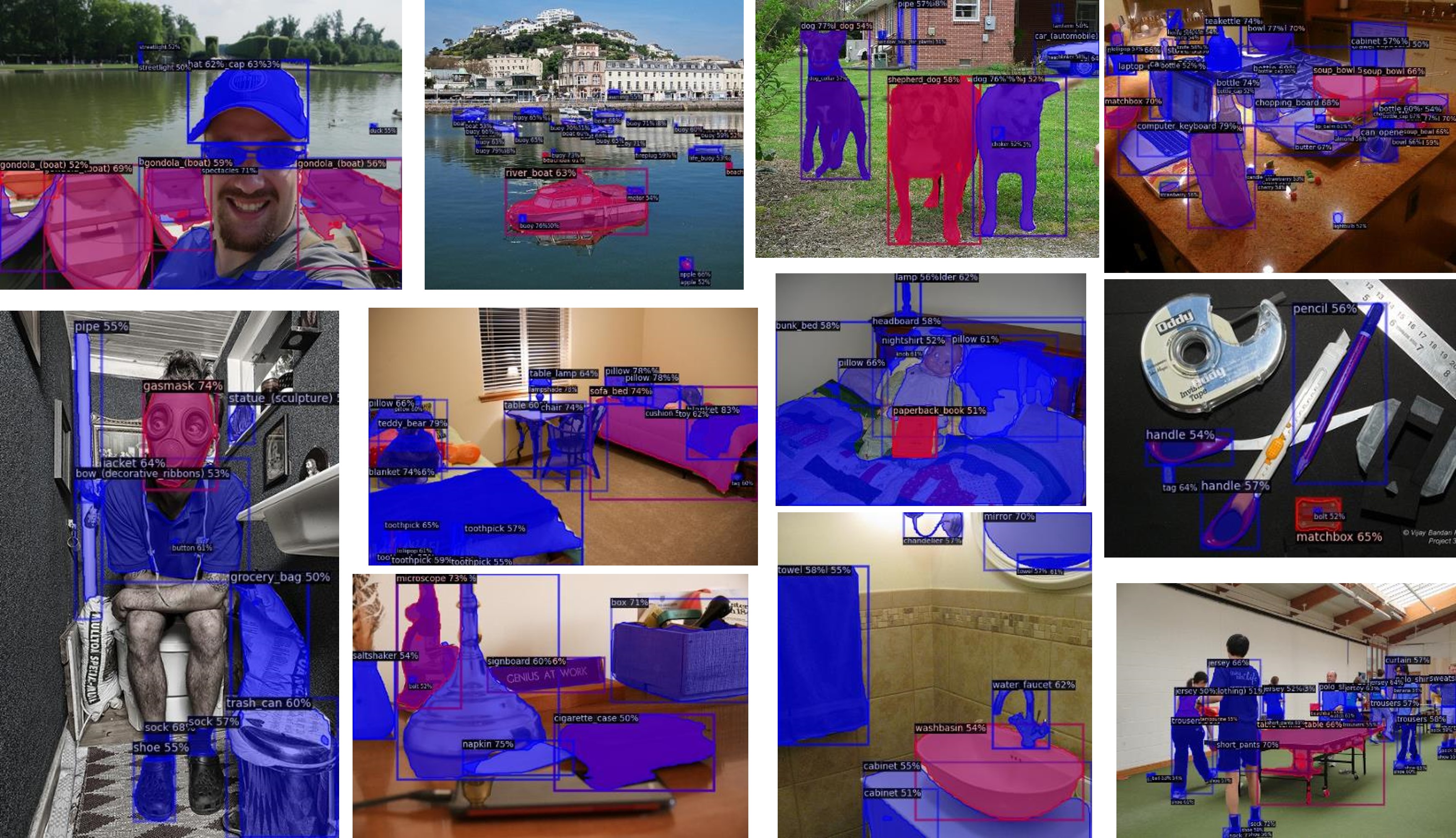}
    \caption{Visualization of detection results on OV-LVIS. Red boxes and masks are for novel (rare) categories.}
    \label{fig:lvis_dets}
\end{minipage}
\end{figure*}

 In Figure~\ref{fig:coco_dets} and Figure~\ref{fig:lvis_dets}, we show some detection results of our models trained on COCO and LVIS, respectively.

\section{Acknowledgement}
This research is supported by the National Research Foundation, Singapore under its AI Singapore Programme (AISG Award No: AISG3-PhD-2023-08-048T), the RIE2020 Industry Alignment Fund – Industry Collaboration Projects (IAF-ICP) Funding Initiative, as well as cash and in-kind contribution from the industry partner(s). Besides, we thank Dr. Xiangtai Li for his help in building detectors on CLIP models, \eg reproducing F-VLM.

\bibliography{aaai24}

\end{document}